%% file: causalToM_preprint.tex
\documentclass[12pt]{article}
\usepackage[T1]{fontenc}
\usepackage[margin=1in]{geometry}
\usepackage{graphicx}
\usepackage{amsmath}
\usepackage{amssymb}
\usepackage{xcolor}
\usepackage{tikz}
\usetikzlibrary{arrows.meta, positioning}
\usepackage{hyperref}
\usepackage{booktabs}
\usepackage{array}

\tikzset{
  causal/.style   = {-{Stealth[length=5pt]}, shorten >=2pt,
                     color=edgegray, line width=0.7pt},
  enabling/.style = {-{Stealth[length=5pt]}, shorten >=2pt,
                     color=enabcolor, line width=0.9pt,
                     dashed, dash pattern=on 4pt off 3pt},
  repeated/.style = {-{Stealth[length=5pt]}, shorten >=2pt,
                     color=repcolor, line width=0.9pt,
                     dotted}
}

\title{A Causal Model of Theory of Mind in Conflict for Artificial Intelligence}
\author{Nikolos Gurney\\
  \small Institute for Creative Technologies, University of Southern California\\
  \small 12015 Waterfront Drive, Playa Vista, CA 90094, USA\\
  \small \href{mailto:gurney@ict.usc.edu}{gurney@ict.usc.edu}\\
  \small ORCID: 0000-0003-3479-2037
}
\date{}

\begin{document}

\maketitle

\begin{abstract}
Theory of mind (ToM), the capacity to ascribe mental states to others and 
use those ascriptions for prediction and inference, is widely assumed to be 
essential for effective human-machine integration. Existing AI-ToM models 
address \emph{how} to mentalize, but leave the question of \emph{when} 
largely unaddressed. The central question is: under what situational and 
agent-level conditions is ToM engagement causally warranted in conflict? 
This paper presents a structural causal model formalized as a directed 
acyclic graph (DAG), treating ToM as a mechanism activated by situational and 
agent-level conditions rather than as an always-on capacity. The model 
specifies four exogenous variables capturing situational and agent-level 
conditions, five endogenous mediators, and a mechanistic ToM node producing 
engagement states through three distinct causal pathways: a tractability 
pathway, a reasoning-depth pathway, and an enabling-cause pathway. The 
primary outcome is epistemic accuracy, which decouples social reasoning from 
behavioral policy and generalizes across social phenomena beyond conflict. 
The framework gives AI systems a principled, resource-rational decision 
procedure for mentalizing, with implications for efficiency, trust, and the 
development of robust artificial social intelligence. Simulation validation, 
empirical human-machine teaming studies, and ethical considerations 
arising from conflict-optimized mentalizing are discussed.

\end{abstract}

\section{Introduction}
\label{sec:intro}
Human-machine integration (HMI) and its related concepts promise a future 
enriched by the collaborative efforts of humans and their AI-enabled 
counterparts. Drones will remove soldiers from the front lines, potentially 
reducing loss of human life in warfare~\cite{chen2014human}. Manufacturing 
robots will increase the productivity of factory workers, driving 
lower costs and more consistent quality~\cite{othman2023human}. 
Autonomous vehicles will reduce transit time, improve safety, and 
decongest cities~\cite{aria2016investigation}. These promises hinge on the 
underlying systems being able understand their human counterparts 
\emph{and} anticipate their behavior, account for their beliefs, and 
recognize potential knowledge gaps. These abilities are precisely the 
capabilities theory of mind (ToM) is argued to support. This 
paper seeks to clarify when ToM is causally warranted. 

\subsection{Motivation}
A major challenge for ToM-enabled intelligent systems is knowing when 
to apply ToM reasoning, as it can be computationally expensive and 
frequently unnecessary, even detrimental to a system's goals. Consider the 
DARPA ASIST program, which sought to build AI coaches enabled with ToM-like 
abilities to support search and rescue teams~\cite{freeman2022}. The 
experimental tasks were structured such that participants 
could excel without relying on rich social interaction, making the 
AI helpers' ToM capabilities largely redundant. Although the ASIST agents 
were still able to make valuable contributions~\cite{pynadath2023}, 
deploying ToM can be costly, thus doing so warrants a principled 
account of when it is causally necessary to avoid wasted resources 
and potential performance degradation. 

Simulation work also questions the role of ToM in social reasoning. 
For example, simple finite state machines in the form of coevolving 
automata can coordinate, engage in conflict, and undertake other social 
activities~\cite{miller2022}. If ToM were universally necessary for 
social behavior, machines without it could not produce these activities.
Miller's results force a reckoning of the role ToM and similar capacities 
play in social coordination: if we accept his social framing, then it 
is possible to achieve social behaviors such as conflict, coordination, 
and cooperation without ToM \emph{or} we must accept that the finite 
state machines had higher-order cognitive ability. Likewise, the utility 
of higher-order ToM appears to depend on the complexity of a prediction 
environment~\cite{deweerd2022}. When a prediction environment is complex, 
the ability to enact multiple levels of ToM (i.e., I know that you know 
that I know…) is beneficial. However, when it is not as complex, rich 
ToM reasoning may prove detrimental because it requires computational 
resources and if stopped at the wrong level may result in a false 
conclusion. 

These examples are indicative of the field more generally. After more 
than four decades of continuous research, it still lacks a principled, 
causal account of \emph{when} ToM is worth engaging. 

\subsection{The Open ``When'' Question}
The present work is motivated by three related questions addressing 
the contextual nature of ToM. First, is ToM necessary only when no 
analytical solution exists, e.g., when game-theoretic solution 
concepts are insufficient or inaccessible? Second, can situational 
complexity alone make ToM sufficiently valuable to justify its cost 
even when analytical solutions exist in principle? Third, what are 
the minimum conditions under which ToM makes a meaningful 
contribution to a sufficiently sophisticated agent's social 
reasoning? The causal model presented here is designed to answer 
these questions formally, establishing the structural conditions 
under which ToM engagement is necessary, sufficient, or merely 
contributory. In other words, it is aimed at addressing the open 
question of ``when'' for mentalizing. 

\subsection{Contributions and Paper Organization}
This paper contributes a theoretical causal account of mentalizing, 
specified for conflict scenarios. The formally specified causal model 
(a) fills the gap left by other mechanistic ToM approaches, and (b) 
provides a blueprint for simulation, empirical, and generalization work. 
A concise account of theory of mind research follows and sets the stage for 
these contributions. 

\section{Background and Related Work}
\label{sec:background}
Ascribing mental states to others, particularly in explaining an 
observation about them, is commonly known as theory of mind (ToM)
~\cite{premack1978}. ToM is believed to play a pivotal role across human 
social interactions, such as cooperating and competing. Existing 
explanations of human ToM tend to focus on how it manifests. For example, 
one posits that children function as lay scientists testing social 
theories~\cite{gopnik1992}, and another argues that belief and desire 
concepts (argued to be critical to ToM) are innate thanks to an inferential 
mechanism~\cite{leslie2004}. These explanations trickle into applied 
settings adjacent to human cognition research, e.g., ToM models for 
robots~\cite{gurney2022robots}. Rather than explain ToM mechanistically, 
the proposed model seeks to explain it contextually, i.e., answering 
questions related to the social settings that give rise to it and its 
broader role in social interactions, an approach not taken by previous 
work. 

\subsection{Theory of Mind: Definitions and Scope}
For the purposes of developing a causal model of theory of mind reasoning 
in conflict, this paper uses the terms \emph{theory of mind} and 
\emph{mentalizing} interchangeably. Both terms carry theoretical baggage: 
theory of mind implicitly favors the theory-theory account of social 
cognition~\cite{gopnik1992}, while mentalizing is sometimes reserved for 
the neural and phenomenological dimensions of perspective-
taking~\cite{frith2006neural}. Neither term is neutral. I adopt both for their 
broad recognition while remaining agnostic about the mechanistic account 
of how the capacity operates, a question the present model explicitly 
defers in favor of specifying the causal conditions under which the 
capacity engages~\cite{gurney2022survey}. In general, I use ToM as a 
noun to refer to the ability and mentalizing as a verb to refer to the 
act of doing it. 

False belief tasks are a classic experimental design used in psychology 
to study ToM abilities; a quick review of the Sally-Anne Test will 
facilitate easier placement of the causal approach in the broader 
literature. The Sally-Anne Test came to prominence in the developmental 
psychology literature, specifically as a tool to explore a child's 
developing social cognition~\cite{wimmer83}. In the task, a child watches 
a story with two dolls, Sally and Anne. Sally puts her marble into her 
basket and leaves the room. While she is gone, Anne moves the marble 
into her own box. When Sally returns, the child is asked: ``Where will 
Sally look for her marble'' The empirical assumption is that only 
children with developed theory of mind abilities will be able to 
correctly resolve the puzzle. The test has seen a resurgence in 
popularity thanks to its use in testing the social reasoning abilities 
of large language models~\cite{kosinski24,ullman2023,chen2025theory}.

A component of agent sophistication that warrants brief introduction is 
\emph{game frame recognition}, or the ability to identify which strategic 
structure, if any, best describes the current interaction. An agent with 
high reasoning depth but poor game frame recognition may derive an 
optimal solution to the wrong problem, producing confident but misguided 
behavior. Game frame recognition has received limited formal treatment in 
the ToM literature but is implicit in game-theoretic accounts of social 
cognition~\cite{yoshida2008game} and in cognitive hierarchy models where 
agents are assumed to recognize the relevant strategic structure before 
reasoning about others' depths~\cite{camerer2004cognitive}. In the 
present model it is treated as a component of a more comprehensive agent 
attribute, their sophistication, rather than a standalone variable. 
The reasoning of this principled simplification is discussed in 
Section~\ref{sec:model}.

\subsection{Mechanistic Accounts and Their Limitations}
Existing explanations of human ToM tend to focus on how it manifests. 
Theory-theory, arguably the most popular, posits that children function 
as lay-scientists who develop theories of their social world~
\cite{gopnik1992}. The basic premise of theory-theory is Bayesian in 
nature, so it lends itself to network modeling and related computational 
approaches, such as inverse reinforcement learning~\cite{jara2019theory}. 
Theory-theory does not support direct empirical falsification in humans 
(scientists cannot ethically lesion a child's brain or impoverish their 
social lives). Moreover, other cognitive processes impair our ability to 
conduct valid second-order empirical tests: people are often biased by their 
own knowledge when trying to take the perspective of somebody who is less 
informed. It turns out that this is true for children who are asked to 
complete the Sally-Anne Test: when the ``curse of knowledge'' was removed, 
three-year-olds performed significantly better, much like their older 
counterparts~\cite{ghrear2021}. 

An alternative explanation argues that belief and desire concepts, which 
the approach argues to be critical to ToM, are innate thanks to an 
inferential mechanism \cite{leslie2004}. The inferential mechanism 
perspective and related ``core mechanism'' theories argue that theory-
theory explanations are incomplete. They typically posit some innate 
solution, such as a belief-desire mechanism, that is hardwired into the 
brain. Core mechanism explanations are often criticized using empirical 
evidence that suggests a conceptual change in reasoning during the preschool 
years~\cite{wellman2001meta}.

The theory-theory and core mechanism accounts rely on paradigms that may
not be measuring ToM at all, as seen in critics of tasks like the Sally-
Anne Test~\cite{ghrear2021}. Additionally, there is mounting evidence 
that many social behaviors occur without reliance on higher-order 
processes, an ability referred to as 
submentalizing~\cite{heyes2014,santiesteban2014}. Simulation work 
further undermines the necessity assumption: finite state machines 
achieve social behaviors, including conflict, without anything resembling 
ToM~\cite{miller2022}. These observations collectively suggest that ToM 
is neither universally necessary nor uniformly deployed, but rather 
sensitive to situational conditions, such as the complexity of the 
interaction~\cite{deweerd2022}, the degree of information asymmetry 
between agents~\cite{wimmer83}, and whether an analytical solution to 
the interaction exists~\cite{yoshida2008game}.

\subsection{ToM in Social Interaction: Necessity, Sufficiency, and Submentalizing}
A foundational assumption in ToM research is that the capacity is 
essential to effective social interaction. This assumption deserves 
scrutiny. Interacting automata, or finite state machines that transition 
through a fixed number of states using simple rules, exhibit emergent 
behaviors that achieve the same outcomes as conflict, cooperation, 
coordination, and other social phenomena through coevolution alone~
\cite{miller2022}. Unless one is prepared to attribute 
ToM to finite state machines, ToM is not a necessary condition for 
social behavior in its basic forms.

The picture becomes more nuanced when complexity enters. De Weerd et al. 
demonstrated that higher-order ToM (reasoning about what others think 
you think) is beneficial in unpredictable negotiation environments but 
detrimental in simpler ones, where the computational cost outweighs the 
epistemic gain~\cite{deweerd2022}. This finding directly motivates 
treating ToM as contextually rather than universally engaged: the value 
of mentalizing is a function of situational conditions, not a fixed 
property of the capacity itself.

Submentalizing further complicates the picture. Heyes argues that many 
behaviors attributed to ToM may instead reflect domain-general cognitive 
mechanisms that simulate mentalizing effects without engaging the actual 
process~\cite{heyes2014}. The dot task, in which participants were found to 
respond faster when their perspective aligns with an avatar's, was long 
taken as evidence of automatic perspective-taking. But replacing the avatar 
with a simple arrow produces the same response time advantage in human 
subjects, suggesting the effect may not require mental state attribution 
at all~\cite{santiesteban2014}. Submentalizing is not a failure of ToM but a 
distinct, cheaper process that produces similar behavioral outputs under 
certain conditions.

Together these observations establish that ToM is neither universally 
necessary nor uniformly deployed. They also motivate a design choice in the 
present model: epistemic accuracy rather than behavioral outcome is the 
primary dependent variable. Agents routinely behave in contradiction to 
their epistemic states: a player who knows the Nash equilibrium may defect 
anyway due to emotion or habit~\cite{loewenstein2005hot}; a human operator 
who correctly models an AI system's capabilities may override it due to 
automation bias~\cite{goddard2012automation}. Behavior is an unreliable 
proxy for the quality of social reasoning. Epistemic accuracy is not.

\subsection{ToM in AI and Human-Machine Integration}
The case for ToM in AI systems is well established. Robots endowed 
with ToM communicate more efficiently and are less intrusive on human 
action~\cite{devin2016}. In multi-agent settings, ToM improves 
communication efficiency and knowledge generalization~\cite{wang2021}. 
Deep interpretable ToM models improve performance in collaborative 
tasks~\cite{oguntola2021}. A growing body of survey work documents 
the range of approaches and their applications~
\cite{gurney2022survey,gurney2022robots}.

Existing implementations tend to cluster around a small number of 
architectures. Decision-theoretic approaches model mental states as 
probability distributions updated via Bayesian inference~
\cite{pynadath2005psychsim}. Interactive POMDPs extend this to 
nested belief structures at significant computational cost~
\cite{gmytrasiewicz05}. Bayesian ToM frames mental state inference as 
inverse planning, recovering goals and beliefs from observed behavior~
\cite{baker09}. Each approach answers the how question: given that an agent 
should mentalize, how should it do so?

The when question receives comparatively little attention. The 
DARPA Artificial Social Intelligence in Support of Teams (ASIST) 
program represents the most systematic attempt to deploy ToM-capable AI in 
realistic team settings~\cite{freeman2022}. Evaluations revealed a structural 
problem: the experimental tasks could be completed without rich social 
interaction, leaving ToM-enabled agents with little causal work to 
do~\cite{pynadath2023}. The lesson is not that ToM is unimportant but that 
deploying it without a principled account of when it is causally warranted 
produces systems that are sophisticated in ways that do not matter for the 
task at hand.

No existing AI-ToM framework provides a causal account of when mentalizing 
is necessary, sufficient, or merely contributory. The present work 
addresses that gap directly, offering a structural causal model that 
specifies the situational and agent-level conditions under which 
mentalizing is warranted---and when analytical or heuristic reasoning is 
sufficient.

\section{A Causal Model of Theory of Mind in Conflict}
\label{sec:model}

\subsection{Modeling Philosophy}
Extensive research attempts to answer the question of \emph{how} to 
implement mentalizing in AI systems 
\cite{gurney2022survey,gurney2022robots}. The paired questions of when 
ToM engages and the epistemic consequences of doing so---the core of 
the present model---are largely left untreated. We address them using 
a structural causal model formalized as a directed acyclic graph (DAG), 
in which nodes represent variables and directed edges represent direct 
causal relationships~\cite{pearl2009,halpern2005a,halpern2005b}. 
The acyclicity constraint ensures temporal ordering is respected and 
facilitates representing ToM as a mechanism node: one that captures 
whether the capacity is engaged rather than how it operates 
internally~\cite{gurney2022survey,gurney2022robots}. The model is 
static and dyadic, representing a single interaction between two agents; 
generalizations are left for future work. Its primary outcome is 
epistemic accuracy rather than agent behavior, which reflects the 
reality that other factors beyond epistemic accuracy impact behavioral 
outcomes~\cite{camerer1989curse}. 

Two primary classes of variables haracterize the model: exogenous, 
or the ones the model takes as given, and endogenous, the ones that 
the model generates. Exogenous variables should be thought of as 
situational facts and agent traits that exist before the interaction 
occurs. These are important, but the model makes no attempt at 
explaining them. On the other hand, endogenous variables should be 
thought of as emerging from the interaction. They are causally 
produced by other variables in the DAG. The nature of the conflict, 
such as its complexity, and the agents' relative abilities, i.e., 
their degree of sophistication, are determinants of the interaction. 
Signals generated during the interaction or an agent's assumption 
about their own ability to derive an optimal response are generated 
by the interaction. 

Theory of mind is a response to epistemic uncertainty about the 
content of other minds, a foundational principle the model must 
account for~\cite{frith2006neural,yoshida2008game}. The model 
separates the true epistemic state of another agent from what the 
focal agent believes is true in three principled ways. First, it 
differentiates between the reality and an agent's perception of 
how tractable the conflict is. Does an optimal conflict behavior 
exist? Does the agent think it exists? Agent capacity is similarly 
treated: the model assumes the focal agent's capacity but 
generates an estimate of the target agent from the focal agent's 
perspective. Does the focal agent believe they are more or less 
sophisticated than their competitor? Finally, it affords agents 
the opportunity to reject their own ToM reasoning, effectively 
allowing other systems to intervene on a potentially defunct 
model. Does the ToM output seem plausible given the context?

\subsection{Exogenous Variables}
The model takes four distinct inputs: three situational, and one agent-
level. This minimal set of exogenous variables (nodes) provides sufficient 
information to model agents in conflict scenarios while ensuring parsimony. 
The situational variables serve to capture the context of the scenario, 
while the agent-level variable deliberately collapses agent-specific 
measures into a composite that carries the modeling-relevant information. 

\textbf{Conflict Complexity} (C) is a continuous variable on the unit 
interval that reflects the number of agents, interaction stakes, and 
history of the scenario. Simple interactions, those near 0, generally 
do not necessitate ToM \cite{miller2022}, while more complex, those 
near 1, do \cite{deweerd2022}. \textbf{Information Asymmetry} (IA) 
captures the degree to which agents hold different information about 
the conflict, its participants, and their intentions. It also a 
continuous variable on the unit interval. Critically, it serves a vital 
roles as an enabling cause for ToM: without IA, there is nothing to 
mentalize about and ToM engagement as no useful causal work to do 
regardless of other conditions. Whether an agent correctly recognizes 
the degree of asymmetry they face is a separate question from whether 
it exists and a distinction the model handles downstream. The final 
situational variable, \textbf{Objective Tractability} (OT), is binary 
and indicates whether their is a closed-form analytical solution to 
the conflict scenario, such as a Nash equilibrium in the case of a 
interaction that can be modeled using the prisoner's dilemma game~
\cite{yoshida2008game}.

The solitary agent-level exogenous variable, \textbf{Sophistication} 
(S), is a composite of an agent's capacity for recursive reasoning, 
game frame recognition (the ability to identify which, if any, 
strategic structure best describes the current interaction), and 
accurate opponent modeling. S takes a value on the unit interval and 
is fixed within a single interaction, an assumption similar to 
treating ability as constant within an experimental session. Collapsing 
agent capacity to a single parameter is a principled simplification 
consistent with the model's primary aim of guiding AI systems rather 
than modeling human cognition in detail. The principled simplification 
is driven by the assumption that, for algorithmic agents, having a 
higher S is positively correlated with higher likelihood of recognizing 
the correct game frame. That being said, there is some empirical basis 
for this collapse~\cite{camerer2004cognitive}. Explicit modeling of 
how relevant agent traits are projected onto the interval, or making S 
entirely endogenous, is left for future work. 

\subsection{Endogenous Variables}
Five endogenous variables mediate between the exogenous inputs and theory of 
mind engagement, thus separating objective ground truth from the focal agent's 
perspective. This approach ensures that the endogenous variables reflect 
the epistemic territory that ToM operates in: agents form beliefs about the 
state of their world based on knowledge and inference, often resulting in 
agent perception not aligning with ground truth. Each variable (node)  
accomplishes distinct causal work that, if collapsed, would result in the 
model being less informative. 

\textbf{Observable Signals} (OS) are the behavioral and situational 
cues generated during a conflict interaction that carry information 
about an opponent's reasoning capacity and intentions. OS is the sole 
mid-interaction update mechanism in the model, making it a causal parent 
of both perceived tractability and perceived opponent sophistication. 
It also serves as the dyadic coupling mechanism: each agent generates 
signals that feed into the other's OS node, linking the two parallel model 
instances without requiring a full repeated-game structure 
\cite{clark1991grounding}. In effect, OS can be understood as the 
information channel from the environment to the agent beliefs. 

Two variables, \textbf{Perceived Opponent Sophistication} (PS) and 
\textbf{Relative Sophistication} (RS), capture the agent's epistemic 
access to their opponent's capacity. Because sophistication is not 
directly observable, agents form an initial estimate of PS by projecting 
from their own S, with self-projection as the empirically documented 
default~\cite{epley2004perspective}. Observable signals then update PS as 
the interaction unfolds. RS, defined as the ratio S/PS, captures the degree 
and direction of miscalibration: values above 1 indicate the agent believes 
they outmatch the opponent, below 1 the reverse, and $\text{RS}=1$ indicates 
accurate calibration. When miscalibration is sufficiently large in either 
direction, RS drives ToM engagement. In a sense, RS serves as an 
interpretable trigger for agents.

Tractability enters the agent's model in two ways. \textbf{Perceived 
Objective Tractability} (POT) captures the agent's ongoing belief that an 
analytical solution exists. It is anchored on OT but can diverge from it: 
C degrades POT while high S partially counteracts this, and OS updates POT 
as the interaction unfolds \cite{yoshida2008game}. This allows meaningful 
false positives and false negatives, as an agent may believe no solution 
exists when one does, triggering unnecessary mentalizing, or believe one 
exists when it does not, suppressing ToM engagement entirely. Where POT 
captures belief about whether a solution exists, \textbf{Accessible 
Tractability} (AT) captures whether the agent, given their belief about 
the solution's existence (POT), their capacity (S), and the demands 
imposed by C, can actually derive it. As AT decreases, the pressure to 
engage in mentalizing increases.

\subsection{Theory of Mind Engagement}
An ordinal mechanistic variable represents ToM in the causal model, 
meaning that the underlying model is assumed and only its output 
state, $\{0,1,2\}$ is modeled. The ToM output is produced in two stages 
via binary processes: an engagement trigger (not engaged vs. engaged) 
and an acceptance decision (rejected vs. accepted):
\begin{quote}
\noindent Is the engagement threshold met?\\
\hspace*{1em}No \quad$\rightarrow$ ToM = 0: not engaged\\
\hspace*{1em}Yes \quad$\rightarrow$ Is the acceptance threshold met?\\
\hspace*{3em}No \quad$\rightarrow$ ToM = 1: engaged, rejected\\
\hspace*{3em}Yes \quad$\rightarrow$ ToM = 2: engaged, accepted
\end{quote}

The first question, \textit{is the engagement threshold met}, is 
answered by a weighted combination of IA, low AT, and miscalibrated RS 
that exceeds an engagement threshold. This combination creates three 
causal pathways: 
\begin{enumerate}
    \item Tractability pathway: when AT is low (agent cannot derive a
    solution) and/or POT is low (agent doubts a solution exists) 
    \item Reasoning-depth pathway: when |$\text{RS} - 1$| is large,
    (the agent believes they meaningfully over- or under-match the 
    opponent)
    \item Enabling-cause pathway: IA is a necessary precondition, 
    meaning without information asymmetry ToM has no useful causal 
    work to do
\end{enumerate}

The second questions, \textit{is the acceptance threshold met}, is 
answered by evaluating the mentalizing output against signal strength 
(OS), their own abilities (S), and the nature of the conflict (C). 
This functions as an opportunity for outside systems to intervene, 
particularly when there is precedent for not trusting the ToM system, 
meaning other reasoning modes (analytical, intuitive) can override the 
mentalizing output. Notably, once these conditions align, mentalizing is 
difficult to suppress, which the overdetermined pathway structure 
is designed to reflect~\cite{heyes2014,santiesteban2014}.

The model treats conflict scenarios as epistemic games with each 
round of reasoning being a round of the game \cite{yoshida2008game}. 
There are two termination conditions: an agent exhausts their capacity 
(maximal S) or the cost of another round of reasoning is higher than 
the expected increase in payoff. 

\subsection{Epistemic Accuracy and Conflict Behavior}
\textbf{Epistemic Accuracy} (EA) is the weighted output of three 
reasoning modes (analytical, intuitive, and mentalizing) whose mixture 
weights are determined by the ToM engagement state. The analytical 
mode is based on AT and has the most weight when ToM is not engaged 
and AT is high. Intuitive captures heuristic, affect, and similar 
process-based reasoning. It is assumed to always be present at some 
level, but treated as a stochastic placeholder in the current model. 
The intuitive mode is further impacted by rejected mentalizing 
output to capture the effect of an agent investing cognitive effort 
in mentalizing then discarding it and being in a different epistemic 
state than an agent who never engage ToM. Moreover, the intuitive 
mode is meant to capture submentalizing, or domain-general mechanisms 
that simulate mentalizng effects without engaging the actual process, 
as a limiting case \cite{heyes2014}. Mentalizing is ToM-based with a 
non-zero value whenever ToM is engaged and accepted. 

\textbf{Conflict Behavior} (CB) is modeled as the joint product of 
EA and RS. This approach ensures that epistemic states are separated 
from behaviors, as it is common for agents, but especially humans, to 
routinely act contrary to what they believe. Separating EA from CB 
reflects the well-documented unreliability of behavior as a proxy for 
reasoning quality, for example, emotion-driven escalation despite 
knowing the dominant strategy \cite{loewenstein2005hot}, automation 
bias in HMI \cite{goddard2012automation}, and curse-of-knowledge errors 
in false-belief tasks even after passing them \cite{birch2007curse}.
Modeling CB as the joint outcome of EA and RS also sets up an 
important dynamic: CB is driven by RS alone when ToM is not engaged 
which establishes submentalizing as the behavioral limiting case. 

Epistemic accuracy is arguably a cleaner optimization target for AI 
systems than CB. The CB as a combination of EA and RS approach 
effectively decouples the cognitive model from behavioral policy, 
allowing each to be optimized independently. Moreover, EA generalizes. 
Although the present model targets conflict, the epistemic question 
at its core (does the focal agent accurately model the target agent?) 
is stable across social phenomena such as coordination and cooperation  
where behavioral payoffs differ substantially. 

\subsection{The DAG: Full Variable List and Edge Set}
Gathering the variables and their connecting edges will facilitate a 
clean presentation of the model's formalized structural equations. 
Table \ref{tab:variables} covers the four exogenous, five endogenous, 
mechanistic, and outcome variables. Figure \ref{fig:dag}  visualizes 
the variables as nodes with their relevant edges, which are laid 
our in detail below. 

\input{fig_dag}

\begin{table}[t]
\centering
\caption{Model variables, their types, ranges, and 
roles.}\label{tab:variables}
\begin{tabular}{>{\raggedright\arraybackslash}p{3.5cm}
                l
                >{\centering\arraybackslash}p{1.5cm}
                >{\raggedright\arraybackslash}p{4.0cm}}
\hline
Variable & Type & Range & Role \\
\hline
Conflict Complexity (C)
  & Exogenous & $[0,1]$ 
  & Situational difficulty of the conflict \\[6pt]
Information Asymmetry (IA)
  & Exogenous & $[0,1]$ 
  & Enabling cause for ToM engagement \\[6pt]
Objective Tractability (OT)
  & Exogenous & $\{0,1\}$ 
  & Ground truth: does a solution exist? \\[6pt]
Sophistication (S)
  & Exogenous & $[0,1]$ 
  & Agent reasoning and modeling capacity \\[6pt]
Observable Signals (OS)
  & Endogenous & $[0,1]$ 
  & Mid-interaction belief update mechanism \\[6pt]
Perceived Objective Tractability (POT)
  & Endogenous & $[0,1]$ 
  & Agent belief: does a solution exist? \\[6pt]
Perceived Opponent Sophistication (PS)
  & Endogenous & $[0,1]$ 
  & Opponent capacity estimate \\[6pt]
Relative Sophistication (RS)
  & Endogenous & $\mathbb{R}^+$ 
  & Miscalibration ratio $S/PS$ \\[6pt]
Accessible Tractability (AT)
  & Endogenous & $[0,1]$ 
  & Can the agent derive the solution? \\[6pt]
ToM Engagement (ToM)
  & Mechanism & $\{0,1,2\}$ 
  & Engagement state of mentalizing \\[6pt]
Epistemic Accuracy (EA)
  & Outcome & $[0,1]$ 
  & Accuracy of opponent model \\[6pt]
\hline
\end{tabular}
\end{table}

The complete edge set along with the structural equations 
constitute the full model specification. Together, they 
allow the model to make falsifiable, casual predictions about 
the engagement of ToM and the associated epistemic accuracy 
that an agent achieves. Conflict behavior is downstream a 
downstream consequence of the model's outcome (EA), thus 
not part of the DAG. The repeated game extension is represented 
by a dashed line in the edge set and in Figure \ref{fig:dag}, 
but that generalization is left for future work. 

\noindent\textbf{Situational structure:}
\begin{gather*}
C \rightarrow IA, \quad C \rightarrow OT, \quad IA \rightarrow OT \\
C \rightarrow OS, \quad IA \rightarrow OS
\end{gather*}

\noindent\textbf{Agent and tractability:}
\begin{gather*}
S \rightarrow POT, \quad OT \rightarrow POT, \quad
OS \rightarrow POT, \quad C \rightarrow POT \\
POT \rightarrow AT, \quad S \rightarrow AT, \quad C \rightarrow AT \\
S \rightarrow PS, \quad OS \rightarrow PS \\
S \rightarrow RS, \quad PS \rightarrow RS
\end{gather*}

\noindent\textbf{ToM engagement:}
\begin{gather*}
AT \rightarrow ToM, \quad POT \rightarrow ToM
  \quad \text{\emph{(tractability pathway)}} \\
RS \rightarrow ToM
  \quad \text{\emph{(reasoning-depth pathway)}} \\
IA \dashrightarrow ToM
  \quad \text{\emph{(enabling-cause pathway)}}
\end{gather*}

\noindent\textbf{Outcome:}
\begin{gather*}
ToM \rightarrow EA, \quad RS \rightarrow EA, \quad AT \rightarrow EA
\end{gather*}

\noindent\textbf{Repeated-game extension:}
\begin{gather*}
EA \dashrightarrow OS
  \quad \text{\emph{(deferred to future work)}}
\end{gather*}


\subsection{Structural Equations}
\label{sec:equations}
The structural equations below specify the model at the 
computational level of analysis~\cite{marr82}, establishing 
the causal relationships and directional effects among 
variables without committing to specific functional forms. 
Functional form specification, including the choice of nonlinear 
versus linear relationships, parameter estimation, and node 
measurement scales, is explicitly deferred to the simulation 
phase described in Section~\ref{sec:future}, where empirical 
calibration becomes tractable.

\paragraph{Observable signals.}
\begin{equation}
  OS = f_1(C,\, IA) + \varepsilon_1
\end{equation}
Increasing in both $C$ and $IA$ with the functional form 
left unspecified; a nonlinear form is likely appropriate given 
that high complexity generates richer but noisier signals. 

\paragraph{Perceived objective tractability.}
\begin{equation}
  POT = \sigma\!\left(OT + \alpha \cdot OS - \beta \cdot C
        + \gamma \cdot S\right) + \varepsilon_2
\end{equation}
$OT$ is the ground truth anchor. $OS$ updates perception upward, 
$C$ degrades it, and $S$ improves calibration. $\sigma$ denotes 
the logistic sigmoid, transforming output to $[0,1]$. 

\paragraph{Perceived opponent sophistication.}
\begin{equation}
  PS = \delta \cdot S + (1 - \delta) \cdot g(OS) + \varepsilon_3,
  \qquad \delta \in [0,1]
\end{equation}
$\delta$ is a fixed anchoring weight. High $\delta$ reflects strong 
self-projection while low $\delta$ reflects high responsiveness to 
observable signals. $g(\cdot)$ is a monotonic function mapping 
signals to a sophistication estimate.

\paragraph{Relative sophistication.}
\begin{equation}
  RS = \frac{S}{PS} + \varepsilon_4
\end{equation}
RS values above 1 indicate overestimation of advantage, below 1 
indicate underestimation, and $RS = 1$ indicates accurate calibration. 
The ratio formulation was chosen instead of a simple difference for 
interpretability across varying sophistication ranges.

\paragraph{Accessible tractability.}
\begin{equation}
  AT = \sigma\!\left(\beta_1 \cdot POT - \beta_2 \cdot C
       + \beta_3 \cdot S\right) + \varepsilon_5
\end{equation}
AT is increasing in $POT$ and $S$, decreasing in $C$. $S$ enters as 
capacity available, not capacity deployed.

\paragraph{Theory of mind engagement.}
ToM engagement is a two-stage process producing a three-state 
variable $ToM \in \{0, 1, 2\}$.

\noindent\textit{Stage 1 --- Engagement:}
\begin{equation}
  E = \mathbf{1}\!\left[
        \lambda_1 \cdot IA
        + \lambda_2 (1 - AT)
        + \lambda_3 \cdot |RS - 1|
        > \theta_E
      \right]
\end{equation}
$IA$ acts as an enabling cause, entered additively. $AT$ enters 
inversely: accessible solutions reduce mentalizing pressure. 
$|RS - 1|$ captures that large mismatches in either direction 
increase engagement pressure. $\theta_E$ is treated as fixed for 
initial analyses, but dynamic thresholds are a candidate for 
simulation exploration.

\noindent\textit{Stage 2 --- Acceptance, conditional on engagement:}
\begin{equation}
  A = \mathbf{1}[E = 1] \cdot
      \mathbf{1}\!\left[
        \lambda_4 \cdot OS
        + \lambda_5 \cdot S
        - \lambda_6 \cdot C
        > \theta_A
      \right]
\end{equation}
Conditional on engagement, acceptance is increasing in signal 
strength and sophistication, but decreasing in complexity. $\theta_A$ 
is conceptually distinct from $\theta_E$: the former governs 
situational triggering, hwile the latter governs confidence in the 
mentalizing output.

\noindent\textit{Combined state:}
\begin{equation}
  ToM = E + A
\end{equation}

\paragraph{Epistemic accuracy.}
\begin{equation}
  EA = w_1 \cdot f_{\mathrm{analytical}}(AT)
     + w_2 \cdot f_{\mathrm{ToM}}(RS)
     + w_3 \cdot f_{\mathrm{intuitive}}
     + \varepsilon_6
\end{equation}
where $w_1 + w_2 + w_3 = 1$ and the mixture weights are functions 
of the ToM engagement state:
\begin{itemize}
  \item $ToM = 0$: $w_2 = 0$; epistemic accuracy is driven by 
        analytical and intuitive modes only.
  \item $ToM = 1$: $w_2 = 0$, but rejected mentalizing leaves a 
        trace in the intuitive mode, $w_3$, coloring subsequent reasoning. 
        The functional form of this trace is not specified as part of the 
        causal model. 
  \item $ToM = 2$: full mixture; the mentalizing mode carries
        meaningful weight alongside analytical and intuitive modes. 
\end{itemize}
The intuitive mode $f_{\mathrm{intuitive}}$ is treated as a 
stochastic term capturing heuristic or affect-driven reasoning, 
pending future formalization. 

Equation (6) treats IA as a weighted contributor to the engagement 
threshold, not a strict gate. Changing from an additive to 
multiplicative specification, in which $\text{IA}=0$ entirely 
suppresses engagement, would more precisely formalize the enabling 
cause relationship laid out for IA above. That being said, the 
additive form facilitates empirical flexibility by allowing 
for partial engagement pressure even under low asymmetry, a 
feature consistent with observations that mentalizing-like 
behavior can occur across a range of situational conditions~
\cite{heyes2014}. Simulation and empirical research will help 
resolve which approach is best for a given context. 

The functional forms of OS~(Eq.~1), the reasoning mode 
subcomponents of EA~(Eq.~9), and the signal-to-sophistication 
mapping $g(\cdot)$~(Eq.~3) are intentionally left unspecified. Each 
represents an open empirical question that existing literature 
addresses only directionally. For example, the anchoring and 
adjustment literature establishes that self-projection is the 
default and signals drive updating~\cite{epley2004perspective}, 
but does not quantify the mapping. Similarly, the complexity-
signal relationship is established in direction~
\cite{deweerd2022} but not in form. The trace left in the 
intuitive mode under $\text{ToM}=1$, although psychologically 
motivated~\cite{heyes2014}, also has an unspecified functional 
form. Any formalization of these functional forms would require 
making assumptions, as the current evidence does not support 
their definitive specifications. Their empirical calibration 
via simulation and experimentation is a primary goal of future 
work. 

\section{Implications for Artificial Intelligence and Human-\\Machine Teaming}
\label{sec:implications}

For humans, knowing the epistemic content of other minds is critical 
enough that the capability appears to come online autonomously and 
relatively early in development. Children learn to resolve false belief 
tasks without guidance and long before tasks requiring formal logical 
reasoning~\cite{wimmer83}. For artificial intelligence, social reasoning, 
like mentalizing, is not so fundamental. At present, frontier AI models 
only clearly engage in mentalizing when explicitly requested to
~\cite{kosinski24} and tend to fail at generalizing the ability beyond 
canonical examples~\cite{ullman2023}. Whether current evaluations are 
even measuring ToM in a meaningful sense remains an open question
~\cite{hu2025re}.Theoretically, the causal model of ToM in conflict 
gives AI systems a principled approach to modeling the minds of other 
agents, whether machine, human, or otherwise. Importantly, the formal 
causal account generates testable predictions. Practically, 
resource-rational ToM has the potential to improve trust in agents. 
A growing literature documents that the dynamic human-agent 
interactions where goal states, actions, and outcomes are mediated 
by trust and trust is contingent on accurate mentalizing
~\cite{pynadath2023effectiveness,devin2016}. 

\subsection{Answering the ``When'' Question for AI}

The causal structure serves the purpose of giving AI systems a 
decision procedure~\cite{lieder2020resource}. At its core, the 
process is simple: assess the social situation (C, IA, OT, S, POT, 
AT, and RS), and if the enabling cause plus at least one triggering 
pathway are active, engage mentalizing. If not, fall back to analytical 
or heuristic reasoning. This allows a system to save resources when 
ToM does not have useful causal work to do, as its typical 
instantiations are computationally expensive relative to other processes~
\cite{pynadath2005psychsim,gmytrasiewicz05,baker09}.

The present approach shifts the implementation question from \emph{how} to 
\emph{when}. Decision-theoretic approaches, such as those implemented in 
PsychSim, model agent mental states as probability distributions over 
world states, typically updated via Bayesian inference~
\cite{pynadath2005psychsim}. Interactive POMDPs extend this to nested 
belief strucutres, where each agent maintains a model of the other's 
models, with computational costs scaling sharply with recursion depth~
\cite{gmytrasiewicz05}. Bayesian ToM Frames mental state inference as 
inverse planning, which entails identifying the target agents goal 
and working backward to infer the mental states that support that 
goal, essentially recovering mental states like beliefs from observed 
behavior~\cite{baker09}. Generative AI models, such as LLMs, generally 
do not implement specific ToM models, rather their mentalizing behavior 
is thought to emerge implicitly from pretraining. The lack of a formal 
model makes it challenging to predict, control, or deploy conditionally 
their ToM abilities~\cite{kosinski24,ullman2023}. At the other end of 
the spectrum, logic- and heuristic-based systems approximate social 
reasoning through rule-following or pattern matching without maintaining 
explicit mental state representations~\cite{gurney2022survey}. Across 
these approaches, the question of when to engage ToM, and when to fall 
back, remains unaddressed. 

\subsection{Epistemic Accuracy as an Optimization Target}
The epistemic accuracy outcome (eq. 9) is grounded in a separate 
specification of the mentalizing mechanism, the Theory of Mind Utility 
(ToM-U;~\cite{gurney2026theorymindutilityformal}). ToM-U formalizes the 
process by which a focal 
agent constructs and evaluates candidate models of another agent's 
belief states. ToM-U returns a confidence-weighted epistemic state 
estimate as its primary output, not a behavioral prediction, making 
EA the natural optimization target for any system invoking it. 
The present model takes this output as given and concerns itself with 
the causal conditions under which the utility is engaged and the 
epistemic consequences of that engagement.

AI systems that incorporate ToM typically optimize for behavioral 
prediction~\cite{gurney2022robots}---e.g., solving for whether an 
agent will cooperate, defect, or escalate---rather than for the 
accuracy of the underlying mental state representation. This creates 
a problem for mentalizing, as correctly representing another agent's 
mental state does not always directly predict their behavior or clearly 
map to how an agent should respond to that behavior. In framing the 
outcome of the ToM mechanism as epistemic accuracy (rather than 
behavioral success), the current approach gives a clean loss function 
for learning-based systems. This framing is consistent with a  
separately specified mentalizing mechanism
~\cite{gurney2026theorymindutilityformal}, which 
returns epistemic state estimates as its primary output rather than 
behavioral predictions. Together, these two components, the causal 
model governing when ToM engages and the utility specifying what it 
computes,  effectively decouples the cognitive model 
from an AI system's behavioral policy, which facilitates separate 
behavior optimization and social reasoning.

\subsection{Human-Machine Teaming}

The causal model is intended to enable AI that can reason about 
minds in a manner similar to humans: rely on heuristics 
and logic unless the task necessitates, or at minimum is more 
efficiently completed with, higher-order cognition. The core 
hypothesis is that such an ability will not only limit resource 
waste, but also produce more accurate, dynamic social reasoning. 
Testing this hypothesis will require careful empirical design, 
as many prior attempts have failed because the implemented tests 
did not take seriously the causal nature of social reasoning like 
ToM. As an example, the DARPA ASIST program ran a series of 
studies in which AI helpers were meant to deploy ToM or similar 
logic but the interactions could be, maybe optimally, resolved 
without using it~\cite{freeman2022}. That is, ToM was not necessary. 
Taking the causal nature of ToM seriously means designing tasks 
where information asymmetry and the context complexity jointly 
necessitate mentalizing rather than just providing an opportunity 
for it. Realizing the promise of human-machine teaming depends not 
only on endowing AI with mentalizing capabilities, but on equipping 
it with the contextual judgment to deploy them wisely, which is 
what the causal model is designed to provide. 

\subsection{Generalizability Beyond Conflict}
The DAG structure is intentionally modular, making the conflict 
model a natural backbone for generalization across social phenomena. 
Coordination and cooperation share much of the structure with 
different edge weights; commerce and communication will likely 
require extending the model further. Once a set of basic social 
phenomena are modeled independently, a unifying framework can be 
distilled from them and validated through simulation and empirical 
HMI testing across contexts.

\subsection{Ethical Considerations}
The prospect of ToM-capable AI raises ethical considerations that 
warrant serious treatment. However, a responsible analysis requires 
engaging with contested questions about autonomy, agency, and the 
appropriate scope of machine social reasoning that extend well beyond 
the theoretical contribution of the present work. Three identifiably 
priorities for future ethical analysis as this modeling paradigm moves 
forward are: the dual-use potential of conflict-optimized mentalizing 
in adversarial settings, the risk of systematic miscalibration in 
cross-cultural deployments given the anchoring dynamics in the PS 
equation, and the implications of AI systems that selectively withhold 
social reasoning from human interaction partners. Substantive treatment 
of these questions warrants work explicitly designed to address them, 
consistent with the view that ethical analysis of AI capabilities is 
most productive when grounded in empirical evidence of actual system 
behavior rather than theoretical speculation.

\section{Limitations and Future Work}
\label{sec:future}
As the first theoretical introduction to a causal model of ToM reasoning, 
the present model accomplishes a great deal. However, it also leaves 
numerous open questions and unresolved modeling details. First, it is 
a static model of a single interaction. Generalizing it to a repeated, 
dynamic version will significantly expand its utility. Relatedly, 
\textbf{Sophistication} is treated as a fixed, exogenous parameter. 
Although this is arguably true for some AI systems, for humans and many 
contemporary AI models, it almost certainly should be endogenous plus 
update during repeated interactions. This also implies defining its
internal structure (reasoning depth, calibration, game frame 
recognition, etc.).

In general, the functional forms of all the structural equations 
capture what is computed, not the details of how the computation 
is carried out. This is perhaps most obvious when considering the  
node measurement scales. From a narrative perspective, it is easy 
enough to justify whether a given node is binary, ordinal, continuous, 
et cetera, however, to go beyond what is computed to parameterization 
of the how will minimally take simulation work. Another significant 
gap created in specifying the model is that of the intuitive 
reasoning mode. Its presence has good psychological merit, but its 
form is unresolved in the literature and the model will benefit 
most from an empirically grounded rather than assumed form. 

\textbf{Information Asymmetry} holds an enabling causal relationship 
with ToM. The cleanest relationship from a theory perspective would 
be to have IA gate ToM multiplicatively, as it would allow for it to 
go to zero. That beings said, an additive relationship is much more 
empirically flexible, ultimately making the final form an open 
empirical question. 

Simulation work can resolve a number of these limitations; general 
validation of the model is planned as an immediate next step. The 
primary goal is to test whether contextual ToM engagement as 
specified by the model results in comparable epistemic accuracy 
as other methods while requiring fewer reasoning cycles. The 
empirical results of this effort will help establish the 
sufficiency hypothesis and aid in the specification of the open 
forms above. 

Importantly, that work will serve as a catalyst for generalizing 
the causal model to other types of social interaction, including 
coordination, cooperation, commerce, and communication. The 
present work establishes the theoretical foundation that makes 
the simulation and future empirical tests meaningful. The theory 
makes principled decisions about experimental setup, selecting 
the right independent and dependent variables, and evaluating 
results feasible.

\section{Conclusion}
\label{sec:conclusion}
Converging evidence from fields as diverse as computational social 
science, developmental psychology, and negotiations suggests that 
theory of mind is contextually, not universally engaged. Social phenomena 
are observed via the emergent behaviors of simple computational agents that 
automatically follow a predetermined sequence of operations
~\cite{miller2022}. Inference tasks once thought to require social 
cognition, like interpreting the gaze of an agent, are equally facilitated 
by simple symbols, like an arrow~\cite{heyes2014,santiesteban2014}. 
Conversely, excessive social cognition can undermine success in 
negotiations~\cite{deweerd2022}. These observations present a direct 
challenge to AI systems that treat mentalizing as an ``always-on'' capacity. 
Historically, intelligent systems have relied on such an approach to
mentalizing rather than trying to replicate the contextually situated nature 
of human ToM. In contrast, contemporary frontier models require prompting~
\cite{chen2025theory}. The DAG presented here gives AI a principled, 
resource-rational decision procedure for determining when mentalizing is 
warranted. In doing so it takes a step toward the spontaneous social 
reasoning that distinguishes robust artificial social intelligence from 
prompt-dependent mentalizing~\cite{gurney2024spontaneous}. Critically, it 
decouples cognition from behavior by elevating epistemic accuracy as the 
outcome variable. This allows a system's behavior to be optimized independent 
of its social reasoning, enabling behavioral policies to be optimized against 
well-specified epistemic targets. Epistemic accuracy is also generalizable in 
ways that ToM anchored to behaviors is not: it readily extends beyond 
conflict to coordination, cooperation, communication, and other social 
phenomena. More broadly, the conceptual approach of modeling ToM as a 
causally bound cognitive process can serve as a new paradigm for modeling 
other ``expensive'' cognitive functions (e.g., causal reasoning, planning) 
that humans engage selectively.

\section*{Acknowledgments}
Research was sponsored by the Army Research Office and was accomplished under
Cooperative Agreement Number W911NF-25-2-0040. The views and conclusions
contained in this document are those of the authors and should not be
interpreted as representing the official policies, either expressed or implied,
of the Army Research Office or the U.S. Government. The U.S. Government is
authorized to reproduce and distribute reprints for Government purposes
notwithstanding any copyright notation herein.

\section*{Competing Interests}
The author has no competing interests to declare that are relevant to the
content of this article.

\bibliographystyle{plain}
\bibliography{references}

\end{document}

%% file: fig_dag.tex
%

%
\definecolor{sitborder}{HTML}{0f6e56}
\definecolor{sitbg}{HTML}{e1f5ee}
\definecolor{agtborder}{HTML}{534ab7}
\definecolor{agtbg}{HTML}{eeedfe}
\definecolor{endborder}{HTML}{ba7517}
\definecolor{endbg}{HTML}{faeeda}
\definecolor{mechborder}{HTML}{993c1d}
\definecolor{mechbg}{HTML}{faece7}
\definecolor{outborder}{HTML}{5f5e5a}
\definecolor{outbg}{HTML}{f1efe8}
\definecolor{edgegray}{HTML}{9e9c93}
\definecolor{enabcolor}{HTML}{ba7517}
\definecolor{repcolor}{HTML}{534ab7}

\begin{figure}[t]
\centering
\begin{tikzpicture}[
  base/.style = {
    draw, rounded corners=4pt,
    font=\small\ttfamily\bfseries,
    text centered,
    minimum width=1.0cm,
    minimum height=0.65cm,
    inner sep=4pt,
    line width=0.8pt
  },
  sit/.style  = {base, fill=sitbg,  draw=sitborder},
  agt/.style  = {base, fill=agtbg,  draw=agtborder},
  end/.style  = {base, fill=endbg,  draw=endborder},
  mech/.style = {base, fill=mechbg, draw=mechborder, minimum width=1.2cm},
  outc/.style = {base, fill=outbg,  draw=outborder},
  potn/.style = {base, fill=endbg,  draw=endborder, minimum width=1.2cm},
]

\node[sit]  (C)   at (1.25, 0)    {C};
\node[sit]  (IA)  at (3.25, 0)    {IA};
\node[sit]  (OT)  at (5.25, 0)    {OT};
\node[agt]  (S)   at (7.75, 0)    {S};

\node[end]  (OS)  at (1.25, -2.0) {OS};
\node[potn] (POT) at (3.75, -2.0) {POT};
\node[end]  (AT)  at (5.75, -2.0) {AT};
\node[end]  (PS)  at (7.75, -2.0) {PS};

\node[end]  (RS)  at (7.25, -3.75) {RS};

\node[mech] (ToM) at (4.50, -4.10) {ToM};

\node[outc] (EA)  at (4.50, -5.75) {EA};


\draw[causal] (C)  -- (IA);
\draw[causal] (C)  to[bend left=18] (OT);
\draw[causal] (IA) -- (OT);
\draw[causal] (C)  -- (OS);
\draw[causal] (IA) -- (OS);

\draw[causal] (OT.south)  -- (POT.north);
\draw[causal] (OS)  -- (POT);
\draw[causal] (S)   to[bend right=18] (POT);
\draw[causal] (C)   to[bend left=10]  (POT);
\draw[causal] (POT) -- (AT);
\draw[causal] (S)   to[bend right=15] (AT.north);
\draw[causal] (C)   to[bend left=10]  (AT);

\draw[causal] (S)  -- (PS);
\draw[causal] (OS) to[bend left=20]  (PS);
\draw[causal] (S)  to[bend right=25] (RS);
\draw[causal] (PS) -- (RS);

\draw[causal]   (AT)  -- (ToM);
\draw[causal]   (POT) to[bend right=12] (ToM);
\draw[causal]   (RS)  to[bend left=10]  (ToM);
\draw[enabling] (IA)  to[bend right=40] (ToM);

\draw[causal] (ToM) -- (EA);
\draw[causal] (AT)  to[bend left=15]  (EA);
\draw[causal] (RS)  to[bend right=12] (EA);

\draw[repeated] (EA.west) to[out=120, in=270, looseness=1.2] (OS.south);

\end{tikzpicture}

\smallskip
\begin{tikzpicture}[
  leg/.style = {draw, rounded corners=3pt,
                font=\footnotesize\ttfamily,
                minimum width=1.4cm, minimum height=0.45cm,
                inner sep=3pt, line width=0.7pt},
  node distance = 0.3cm and 0.35cm
]
  \node[leg, fill=sitbg,  draw=sitborder]                (l1) {Sit.\ exog.};
  \node[leg, fill=agtbg,  draw=agtborder, right=0.35cm of l1] (l2) {Agt.\ exog.};
  \node[leg, fill=endbg,  draw=endborder, right=0.35cm of l2] (l3) {Endogenous};
  \node[leg, fill=mechbg, draw=mechborder,right=0.35cm of l3] (l4) {Mechanism};
  \node[leg, fill=outbg,  draw=outborder, right=0.35cm of l4] (l5) {Outcome};

  \coordinate (ek) at ([yshift=-0.65cm] l1.west);
  \draw[causal]
        (ek) -- ++(0.65cm,0)
        node[right, font=\footnotesize\rmfamily] {Causal edge};
  \draw[enabling]
        ([xshift=2.8cm] ek) -- ++(0.65cm,0)
        node[right, font=\footnotesize\rmfamily] {Enabling cause};
  \draw[repeated]
        ([xshift=6.2cm] ek) -- ++(0.65cm,0)
        node[right, font=\footnotesize\rmfamily] {Repeated game ext.};
\end{tikzpicture}

\caption{Structural causal model of Theory of Mind engagement in
conflict. Solid grey arrows are causal edges. The dashed amber arrow
(IA\,$\rightarrow$\,ToM) is the enabling cause. The dotted purple arc
(EA\,$\rightarrow$\,OS) indicates the repeated-game extension, deferred
to future work. Node shading encodes variable type: teal~=~situational
exogenous (C, IA, OT); purple~=~agent exogenous (S); amber~=~endogenous
(OS, POT, PS, RS, AT); coral~=~mechanism (ToM); grey~=~outcome (EA).}
\label{fig:dag}
\end{figure}
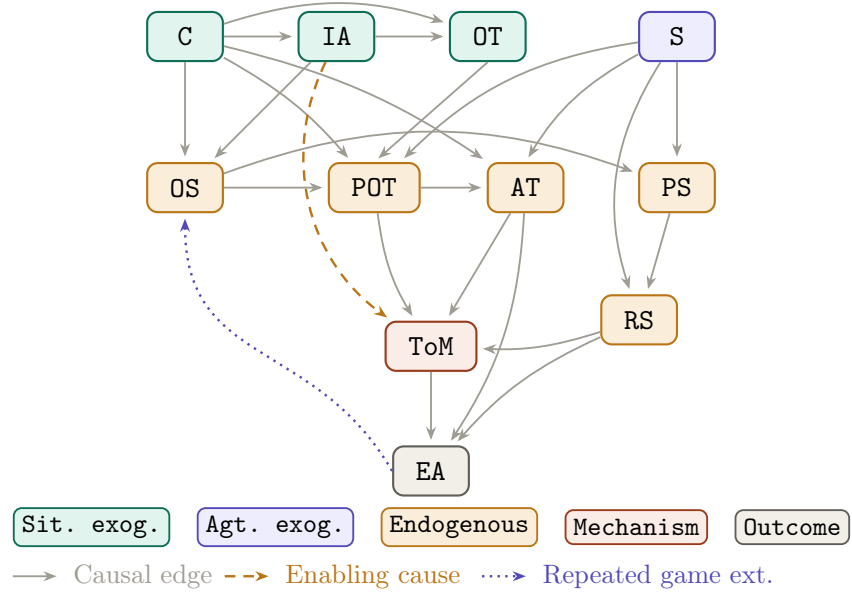